\documentclass{article}
\usepackage{arxiv}

\usepackage[utf8]{inputenc} % allow utf-8 input
\usepackage[T1]{fontenc}    % use 8-bit T1 fonts
\usepackage{hyperref}       % hyperlinks
\usepackage{url}            % simple URL typesetting
\usepackage{booktabs}       % professional-quality tables
\usepackage{amsfonts}       % blackboard math symbols
\usepackage{nicefrac}       % compact symbols for 1/2, etc.
\usepackage{microtype}      % microtypography
\usepackage{graphicx}

\usepackage{amsmath}
\usepackage{amssymb}
\graphicspath{{figures/}}
\usepackage{multirow}

\usepackage{multirow}
\usepackage{booktabs, caption, makecell}
\usepackage{threeparttable}
\usepackage{url}
\usepackage{amsfonts}
\let\vec\mathbf
\title{Multi-Modal Recognition of Worker Activity for Human-Centered Intelligent Manufacturing}

\author{
	Wenjin Tao\\
	Department of Mechanical and Aerospace Engineering\\
	Missouri University of Science and Technology\\
	Rolla, MO 65409, USA \\
	\texttt{w.tao@mst.edu} \\
	\And
	Ming C. Leu \\
	Department of Mechanical and Aerospace Engineering\\
	Missouri University of Science and Technology\\
	Rolla, MO 65409, USA \\
	\texttt{mleu@mst.edu} \\
	\AND
	Zhaozheng Yin \\
	Department of Computer Science\\
	Missouri University of Science and Technology\\
	Rolla, MO 65409, USA \\
	\texttt{yinz@mst.edu} \\
}

\begin{document}
\maketitle

\begin{abstract}
% why it is important
In a human-centered intelligent manufacturing system, sensing and understanding of the worker's activity are the primary tasks.
%quantification and evaluation of the worker's performance, as well as to provide awareness for onsite instructions with augmented reality and human-robot collaboration.
% what do we do
In this paper, we propose a novel multi-modal approach for worker activity recognition by leveraging information from different sensors and in different modalities. 
%Instead of extracting handcrafted features, which is time-consuming and requires domain knowledge, our proposed method digests the raw signals directly.
% how: imu stream
Specifically, a smart armband and a visual camera are applied to capture Inertial Measurement Unit (IMU) signals and videos, respectively. 
For the IMU signals, we design two novel feature transform mechanisms, in both frequency and spatial domains, to assemble the captured IMU signals as images, which allow using convolutional neural networks to learn the most discriminative features.
%Orientation Changing History (OCH) to represent the spatial-domain information.
%The raw 10-channel IMU signals are stacked to form a signal image. This image is transformed into an activity image by applying Discrete Fourier Transformation (DFT) and then fed into a Convolutional Neural Network (CNN) for feature extraction, resulting in a high-level feature vector.
% how: video stream
Along with the above two modalities, we propose two other modalities for the video data, at the video frame and video clip levels, respectively.
Each of the four modalities returns a probability distribution on activity prediction. Then, these probability distributions are fused to output the worker activity classification result.
% result
A worker activity dataset of 6 activities is established, which at present contains 6 common activities in assembly tasks, i.e., grab a tool/part, hammer a nail, use a power-screwdriver, rest arms, turn a screwdriver, and use a wrench. 
The developed multi-modal approach is evaluated on this dataset and achieves recognition accuracies as high as 97\% and 100\% in the leave-one-out and half-half experiments, respectively.
\end{abstract}


\section{Introduction}\label{main}
% Motivation: data -> machine/deep learning -> AI
Industrial big data has been increasingly accessible and affordable, benefiting from the 
availability of low-cost sensors and the development of Internet-of-Things (IoT) technologies~\cite{jeschke2017industrial, lee2015industrial}, which builds up the data foundation for advanced manufacturing. A variety of methods and algorithms have been developed to learn valuable information from the data, and to make the manufacturing more intelligent~\cite{nagorny2017big}.
With the recent fast growing of Artificial Intelligence (AI) technologies, especially deep learning~\cite{lecun2015deep} and reinforcement learning~\cite{kober2013reinforcement} methods, AI boosted manufacturing has been increasingly attractive in both the scientific research and industrial applications.

In an intelligent manufacturing system involving workers, recognition of the worker's activity is one of the primary tasks. It can be used for quantification and evaluation of the worker's performance, as well as to provide onsite instructions with augmented reality.
Also, worker activity recognition is crucial for human-robot interaction and collaboration.
It is essential for developing human-centered intelligent manufacturing systems.

\subsection{Related Work}
% Video based and Wearable sensors based
In the computer vision area, image/video-based human activity recognition using deep learning methods has been intensively studied in recent years and unprecedented progress has been made~\cite{ji20133d, carreira2017quo}. However, visual-based recognition suffers from the occlusion issue, which affects the recognition accuracy.
%However, it needs tremendous labeled data to train the model, which is costly both economically and computationally. 
% Wearable device
Wearable devices, such as an armband embedded with an Inertial Measurement Unit (IMU),
%or surface electromyography (sEMG) sensors, 
directly sense the movement of human body, 
%or the level of muscle activation, 
which can provide information on the body status. In addition, there are a lot of inexpensive wearable devices in the market, such as smart armbands and smartphones, which are widely used in activity recognition tasks. 
Wearable devices are directly attached to the human body and thus do not have the occlusion issue. Nevertheless, a wearable device can only sense the human body activity locally, it is challenging to precisely recognize an activity involving multiple body parts.
Although multiple devices can be applied to simultaneously sense the activity globally, it makes the system cumbersome and brings discomfort to the user.

%Nevertheless, wearable devices loses the full picture of the human body which cannot make a fine-grained details. The resolution may be low for some fine-grained activities. few sensor not precise but more not convenient.
%This paper focuses on worker activity recognition using a Myo armband with IMU and sEMG signals.

% in manufacturing area
Worker activity recognition in the manufacturing area is still an emerging topic and few studies have been made.
%2006
Stiefmeire et al.~\cite{stiefmeier2006combining} utilized ultrasonic and IMU sensors for worker activity recognition in a bicycle maintenance scenario using a Hidden Markov Model classifier.
%2007
Later they proposed a string-matching based segmentation and classification method using multiple IMU sensors for recognizing worker activity in car manufacturing tasks~\cite{stiefmeier2007fusion, stiefmeier2008wearable}.
%2009
Koskimaki et al.~\cite{koskimaki2009activity} used a wrist-worn IMU sensor to capture the arm movement and a K-Nearest Neighbor model to classify five activities for industrial assembly lines.  
%2016
Maekawa et al.~\cite{maekawa2016toward} proposed an unsupervised measurement method for lead time estimation of factory work using signals from a smartwatch with an IMU sensor.
%2018
Recently, deep learning methods have been introduced to recognize worker activity in human-robot collaboration studies~\cite{wang2018deep, petruck2018using}.

%\subsection{Related work}

% feature extraction
In general, the activity recognition task can be broken down into two subtasks: feature extraction and subsequent multiclass classification. 
To extract more discriminative features, various methods have been applied to the raw signals in the time or frequency domain, e.g., mean, correlation, and Principal Component Analysis~\cite{anguita2013public, peterek2014comparison, chang2015hierarchical, ronao2014human}.
% classifiers
Different classifiers have been explored on the features for activity recognition, such as the Support Vector Machine~\cite{anguita2013public, chang2015hierarchical}, Random Forest, K-Nearest Neighbors, Linear Discriminant Analysis~\cite{peterek2014comparison}, and Hidden Markov Model~\cite{ronao2014human}.
To effectively learn the most discriminative features, Jiang et al.~\cite{jiang2015human} proposed a method based on Convolutional Neural Networks (CNN). They assembled the raw IMU signals into an activity image, which enabled the CNN model to automatically learn the discriminative features from the activity image for classification.
%Anguita et al.~\cite{anguita2013public} used a waist-mounted smartphone to capture IMU signals. A set of handcrafted features, e.g. mean, correlation and signal magnitude area, were extracted and then fed into a Support Vector Machine (SVM) classifier to distinguish 6 activities. 

% Transfer learning
%Transfer leaning refers to the situation where what has been learned in one setting is exploited to improve generalization in another setting~\cite{}. It has been applied to many deep learning tasks.

% Multi-modal fusion
%Multi-modal fusion is a good solution to solving a problem from different perspectives and integrate all the modalities together organically to have a comprehensive result. It has shown good results in some studies~\cite{}.

\subsection{Proposed Method}\label{our proposal}
\begin{figure}[h]
\begin{center}
\includegraphics[width=1.0\linewidth]{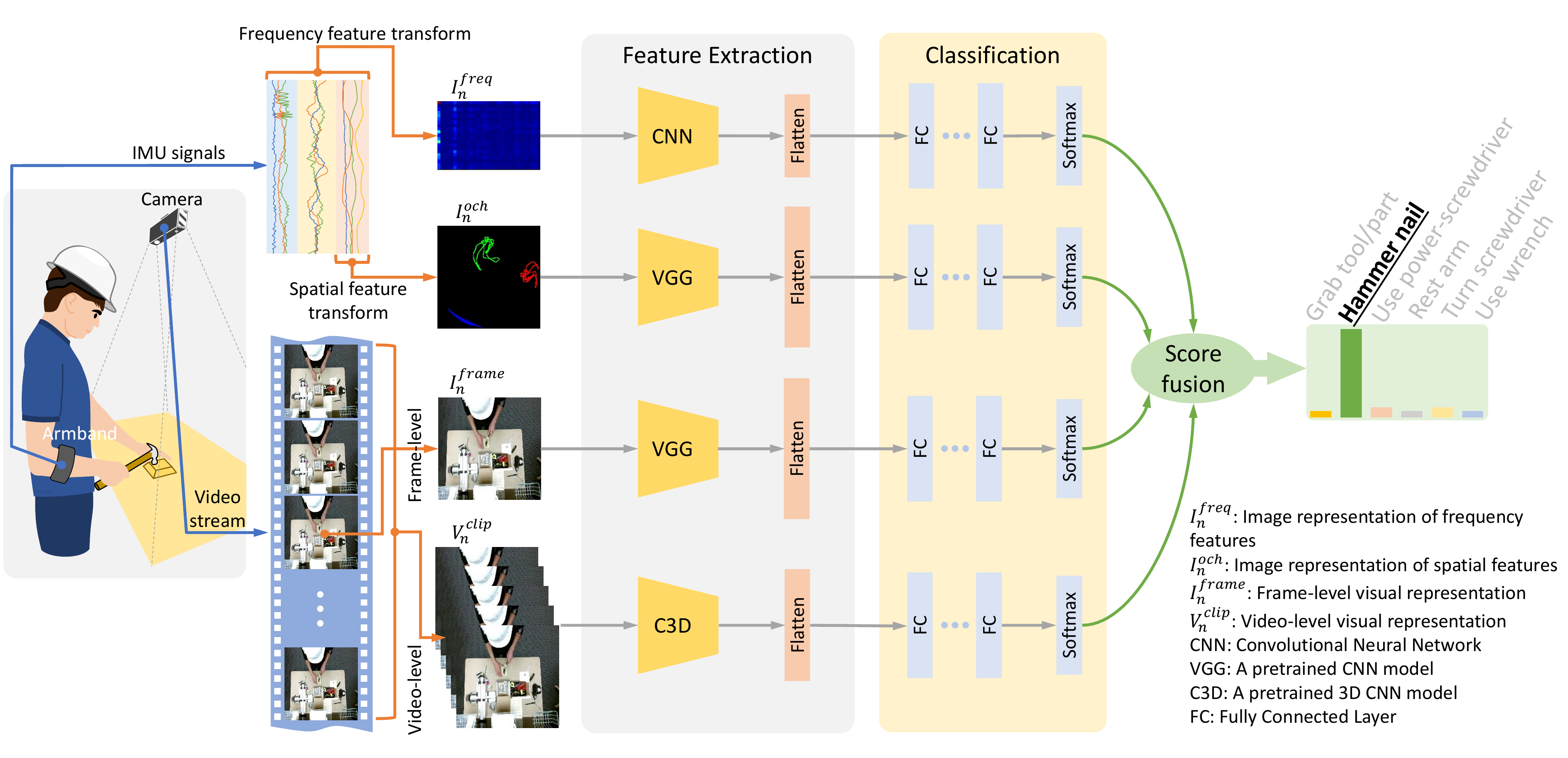}
\end{center}  
\caption{Overview of our multi-modal approach for worker activity recognition.}
%\label{fig:long}
\label{fig:overview_method}
\end{figure}

%drawbacks
Few attempts have been made for the worker activity recognition in the manufacturing field, and most of them only use single sensing modality, which cannot guarantee robust recognition under various circumstances.
In the present research, to comprehensively perceive the worker, we choose a smart armband to acquire the Inertial Measurement Unit (IMU) signals and a visual camera to capture the image sequence of the worker's activity.
%Motivated by the study of Jiang et al.~\cite{jiang2015human}, we stack the raw IMU signals to form a signal image. This image is transformed into an activity image by applying Discrete Fourier Transformation (DFT) and then fed into a CNN for feature extraction, resulting in a high-level feature vector.
%Another feature vector representing the level of muscle activation is calculated from the raw sEMG signals. Then these two vectors are concatenated and used for worker activity classification. 
An overview of our method is illustrated in Figure~\ref{fig:overview_method}.
For the IMU signals, we design two novel mechanisms, in both the frequency and spatial domains, to assemble the captured IMU signals as images. 
The assembled signal representation allows us to use Convolutional Neural Networks to explore the correlation among time-series signals and learn the most discriminative features for worker activity recognition. 
As for the video data, we propose two modalities, at the frame and video-clip levels, respectively.
Overall, we have four modalities in parallel and each of the four modalities can return a probability distribution on the activity recognition. Then these probabilities are fused to output the worker activity classification result.
To evaluate the method, a worker activity dataset containing 6 common activities in assembly tasks is established.

%Depth images contain distance information from the camera plane to the objects in the camera view, where each pixel represents a measured distance. Therefore, it is easier to segment the target object in a depth image than a color image. Thus, this research focuses on recognizing finger spelling signs from depth images as follows:

The main contributions of our work are as follows:
\begin{itemize}
\item We propose a multi-modal approach for the worker activity recognition in manufacturing, using both wearable devices and visual cameras.

\item 
To take advantage of the powerful learning ability of CNN on images, we design two novel mechanisms to produce 2D signal representations of the IMU signals from wearable devices, in both the frequency and spatial domains.

\item
To synthesize more physical-realistic variations in the training dataset, we propose a kinematics-based data augmentation method for the wearable sensor data. It generates more data by spatial rotation and mirroring, in order to augment variations that cannot be achieved using traditional image augmentation methods. 
%It also employs view data augmentation for more effective training and to reduce potential overfitting. 

%\item To solve the interclass similarity issues caused by perspective variations and partial occlusions, we first make predictions for all individual views and then fuse information from them for the final prediction.
\end{itemize}

%\paragraph{Outline} 
The remainder of this paper is organized as follows.
Section~\ref{multimodal_sensing} discusses how we build up the worker activity dataset. 
Section~\ref{signals_preprocessing} focuses on the novel feature representation and data augmentation. Section~\ref{multimodal_arch} describes the details of neural network architectures, training and testing of the multi-modal activity recognition. The experimental setups and results are described in Sections~\ref{experiment} and~\ref{result}, respectively.
Finally, Section~\ref{conclusion} provides the conclusions of this research.

\section{Multi-modal Sensing and Data Acquisition}\label{multimodal_sensing}
% tasks/6 activities
To establish our dataset of worker activity, six activities commonly performed in assembly tasks are chosen, which are: grab a tool/part (GT), hammer a nail (HN), use a power-screwdriver (UP), rest arms (RA), turn a screwdriver (TS), and use a wrench (UW). 
% experiments
%\textbf{WA6 Dataset} 
There are 8 subjects recruited to conduct a set of tasks (listed in Table I) containing the 6 activities. 
\begin{table}[h]\label{tab:tasks}\footnotesize
\begin{center}
\caption{Tasks for collecting worker activity.}
\begin{tabular}{c l c}
\hline
No. & Tasks & Activities\\
\hline
1 & Grab 30 tools/parts from the 3 containers &  GT\\
2 & Hammer 15 nails into the wooden dummy &  HN\\
3 & Tighten 20 screws using a power-screwdriver &  UP\\
4 & Rest arms for about 60 seconds &  RA\\
5 & Tighten 10 nuts using a screwdriver &  TS\\
6 & Tighten 10 nuts using a wrench &  UW\\
\hline
\end{tabular}
\end{center}
\end{table}

\begin{figure}[h]
\begin{center}
\includegraphics[width=0.65\linewidth]{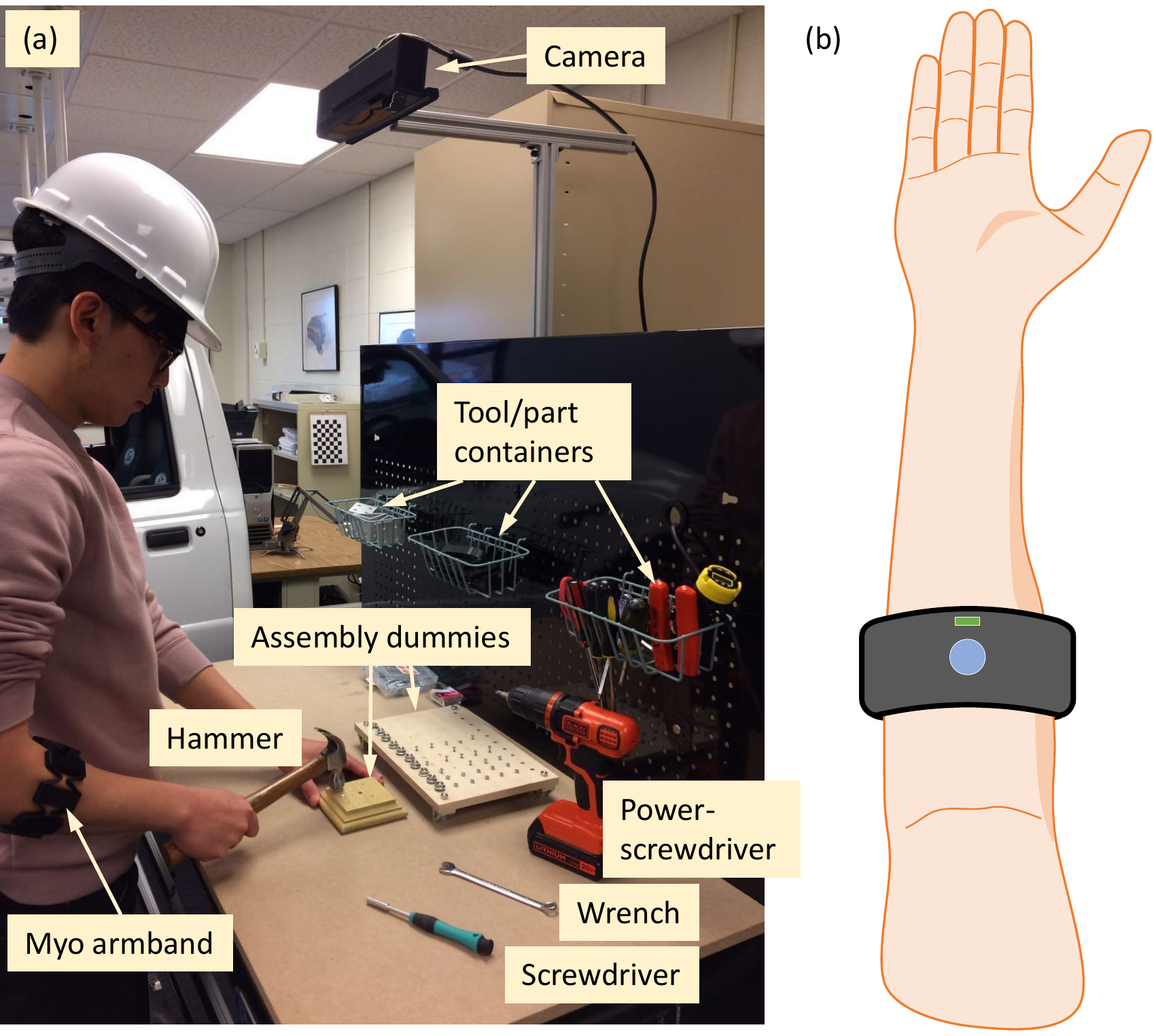}
\end{center}
%\vspace*{-15pt}
   \caption{(a) Data collection setup; (b) Wearing orientation of a right-hand.}
%\label{fig:long}
\label{fig:system_setup}
\end{figure}

As demonstrated in Figure~\ref{fig:system_setup}(a), the subject is asked to stand in front of the workbench, wear a smart armband~\cite{myo} on his/her right forearm with a fixed orientation (Figure~\ref{fig:system_setup}(b)), and perform the tasks on assembly dummies in a natural way. The armband from Thalmic Labs is equipped with IMU sensors for wearable sensor data acquisition. 
% sensors
The IMU returns three types of signals (3-channel acceleration, 3-channel angular velocity, and 4-channel orientation) at the sample rate of 50Hz. 
%A set of 8 sEMG pods attached to the skin return 8 channels of unitless signals in the range of [-128, 127] at the sample rate of 200Hz, which represent the corresponding muscle activations. 
These 10-channel signals captured on a worker are transmitted via Bluetooth to the workstation in real time.

\begin{figure}[h]
\begin{center}
\includegraphics[width=.7\linewidth]{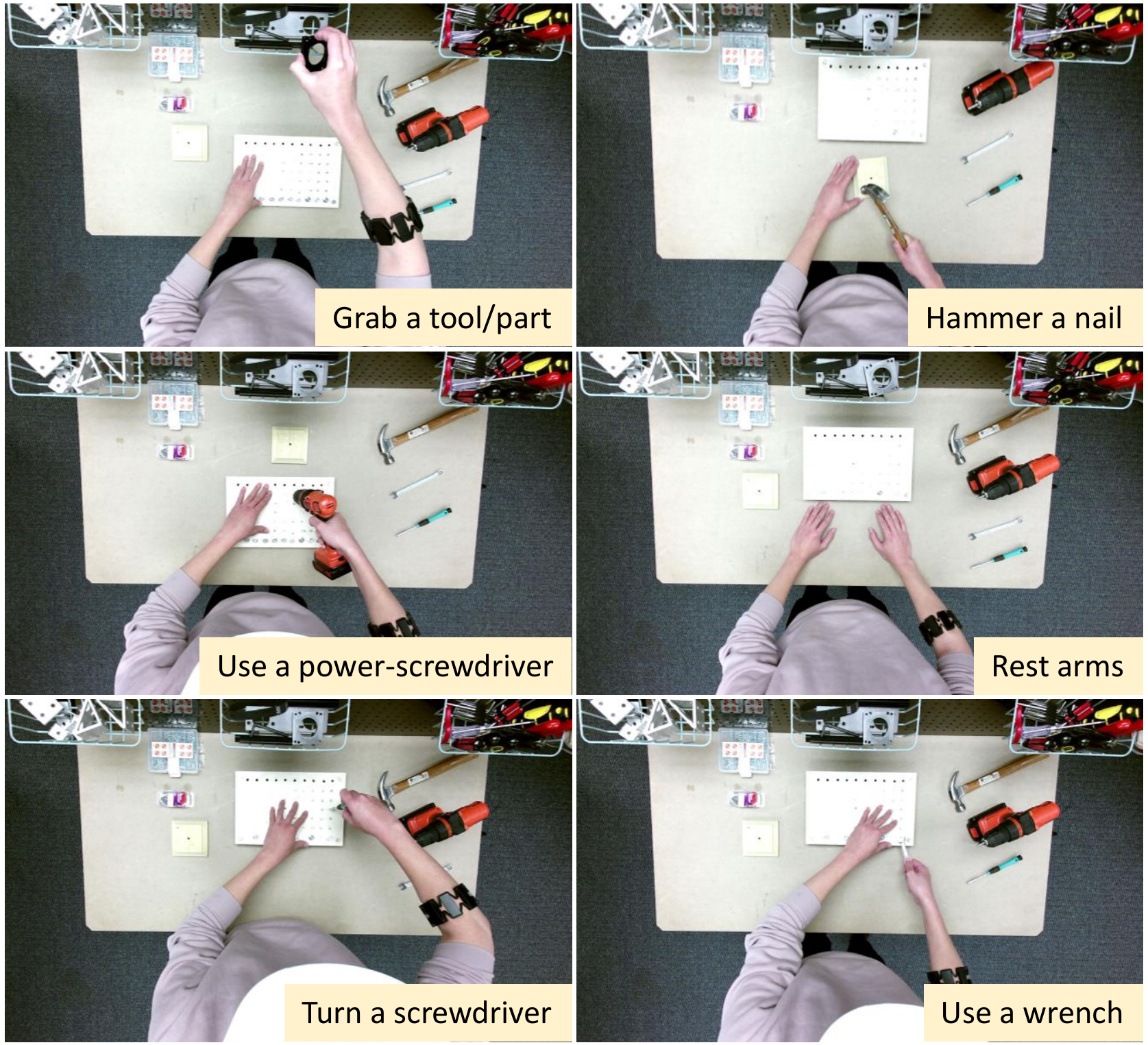}
\end{center}
%\vspace*{-15pt}
   \caption{Examples of the 6 activities captured from the overhung camera.}
%\label{fig:long}
\label{fig:6_activities}
\vspace{0pt}\end{figure}

While collecting wearable sensor data from the armband, an overhung camera is used to record the assembly tasks simultaneously for monitoring the process.  Examples of the 6 activities are shown in Figure~\ref{fig:6_activities}, which are taken from the overhung camera.
% image of 6 activities
% image of dummy

\section{Data Preprocessing, Signal Representation and Data Augmentation}\label{signals_preprocessing}
% segmentation
% sampling
% augmentation
Convolution-based deep learning methods need the input data to be formatted as tensors, for example, with a fixed size of $h\times w\times c$ for images or with a fixed size of $h \times w\times c\times l$  for image sequences (video clips) where $h$, $w$ and $c$ are the height, width and the number of channels of the image, respectively, and $l$ is the image sequence length. Therefore, some preprocessing steps are necessary before the data can be fed into a convolutional neural network. 
In this section we give a detailed description of the pipeline for data preprocessing  and the new methods for signal representation. Furthermore, to generate more realistic data, we propose a kinematics-based augmentation method which is also presented in this section.

\subsection{Data Sampling}\label{sampling}
% segmentation
Although the data (i.e., IMU sensor signals and videos) are collected simultaneously for all tasks and each task consists of only one activity, there still might be some unrelated activities inside the data, such as preparing activities before hammering nails. To address it, the recorded videos are manually annotated to locate the time durations (i.e., the starting and ending timestamps), each of which contains only one of the six activities. These durations are used to segment the raw data (IMU sensor signals and videos).

\begin{figure}[h]
\begin{center}
\includegraphics[width=.6\linewidth]{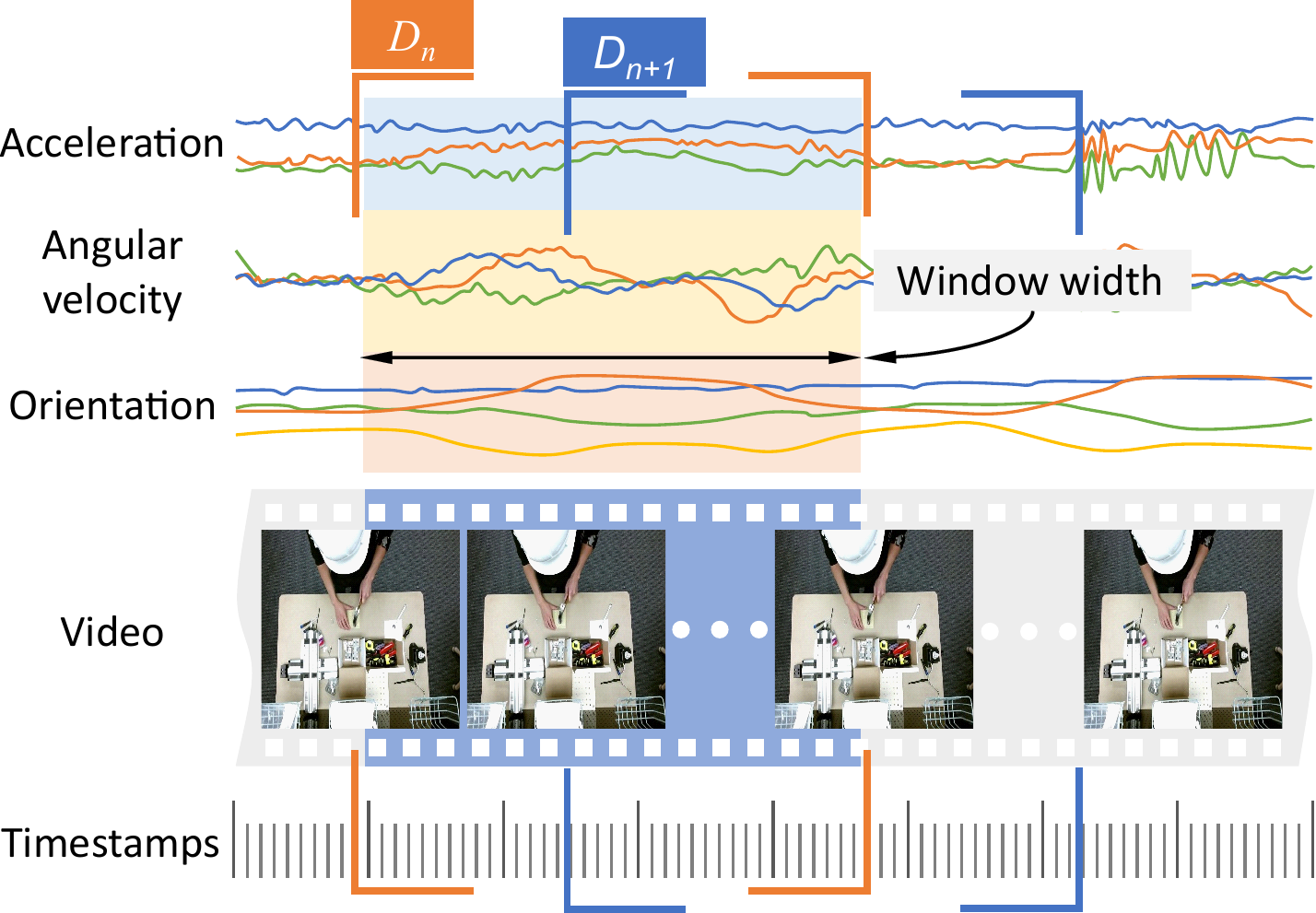}
\end{center}
   \caption{Scheme of the signal sampling method.}
%\label{fig:long}
\label{fig:sampling}
\end{figure}

% sampling %convert the data to the same time length
Usually, the duration of an activity instance ranges from a few seconds to more than one minute. Thus, sampling is needed to prepare the data samples for recognition. 
% imu
As depicted in Figure~\ref{fig:sampling}, the 10-channel IMU signals and the video recording are synchronized with the timestamps. Then the 50Hz IMU signals are sampled using a temporal sliding window with the width of $T=64$ timestamps and 75\% overlap between two windows. Thus, each IMU sample lasts for about 1.3 seconds, which covers at least one activity pattern.
%% emg
%After sampling the IMU signals, the 200Hz sEMG signals are sampled according to the time durations of the IMU samples. Then each sEMG sample has an approximate width of 256 timestamps.
% vid
After sampling the IMU signals, the video recordings are sampled according to the time durations of the IMU samples. Then, each video clip has an approximate length of 38 frames.

After sampling, we denote our dataset as $\mathbb{D}=\{D_1, \cdots, D_n, \cdots, D_N\}$
\begin{equation} \label{eq:Dn}
D_n = \{[s_n, v_n], y_n\}, n\in[1, N]
\end{equation}
where $s_n$ is a sample set of time-series IMU signals, $v_n$ is the corresponding video clip sample,
%can containing different number of frames, 
and $y_n$ is the manually labeled ground truth of the activity class.
%from wiki: A time series is a series of data points indexed in time order. Most commonly, a time series is a sequence taken at successive equally spaced points in time. 
More specifically, $s_n$ a sequence of discrete-time data over $T$ timestamps, $s_n=\{s_{n,1},\cdots,s_{n,t},\cdots,s_{n,T}\}$, and each element is elaborated as
\begin{equation}\label{}
\begin{split}
s_{n,t}^{}=[\underbrace{a_{n,t}^x, a_{n,t}^y, a_{n,t}^z}_{a_{n,t}\text{: acceleration}}, \underbrace{g_{n,t}^x, g_{n,t}^y, g_{n,t}^z}_{g_{n,t}\text{: gyro}}, \underbrace{q_{n,t}^x, q_{n,t}^y, q_{n,t}^z, q_{n,t}^w}_{q_{n,t}\text{: orientation}}], t\in[1, T],
\end{split}
\end{equation}
where ${a}$, ${g}$, and ${q}$ are acceleration, angular velocity, and orientation in quaternion, respectively.

%\subsection{Dataset Analysis}

%We evaluate our method on an established worker activity dataset {WA6}, which has 6 activities performed by 8 subjects. 
After sampling, the quantitative information of the dataset is listed in Table II. There are 11,211 data samples in total. The eight subjects use different amounts of time to finish each task, therefore they have different numbers of data samples for each activity.

\begin{table}[h]\footnotesize
\begin{center}
\caption{Number of data samples for each activity of different subjects.}
\begin{tabular}{c c c c c c c}
\hline
Subject No. & GT & HN & UP & RA & TS & UW\\
\hline
1 & 193 & 140 & 364 & 266 & 222 & 442\\
2 & 302 & 408 & 195 & 56 & 274 & 751\\
3 & 198 & 183 & 171 & 251 & 214 & 567\\
4 & 204 & 172 & 188 & 29 & 82 & 344\\
5 & 187 & 204 & 142 & 43 & 213 & 372\\
6 & 216 & 77 & 179 & 47 & 129 & 301\\
7 & 213 & 196 & 203 & 254 & 231 & 576\\
8 & 200 & 184 & 262 & 145 & 148 & 273\\
\hline
Total & 1713	& 1564 & 1704 & 1091 & 1513 & 3626\\
\hline
\end{tabular}
\end{center}
\label{tab:activity_numbers}
\end{table}

\subsection{Wearable Sensor Signal Representation}
\begin{figure}[h]
\begin{center}
\includegraphics[width=0.9\linewidth]{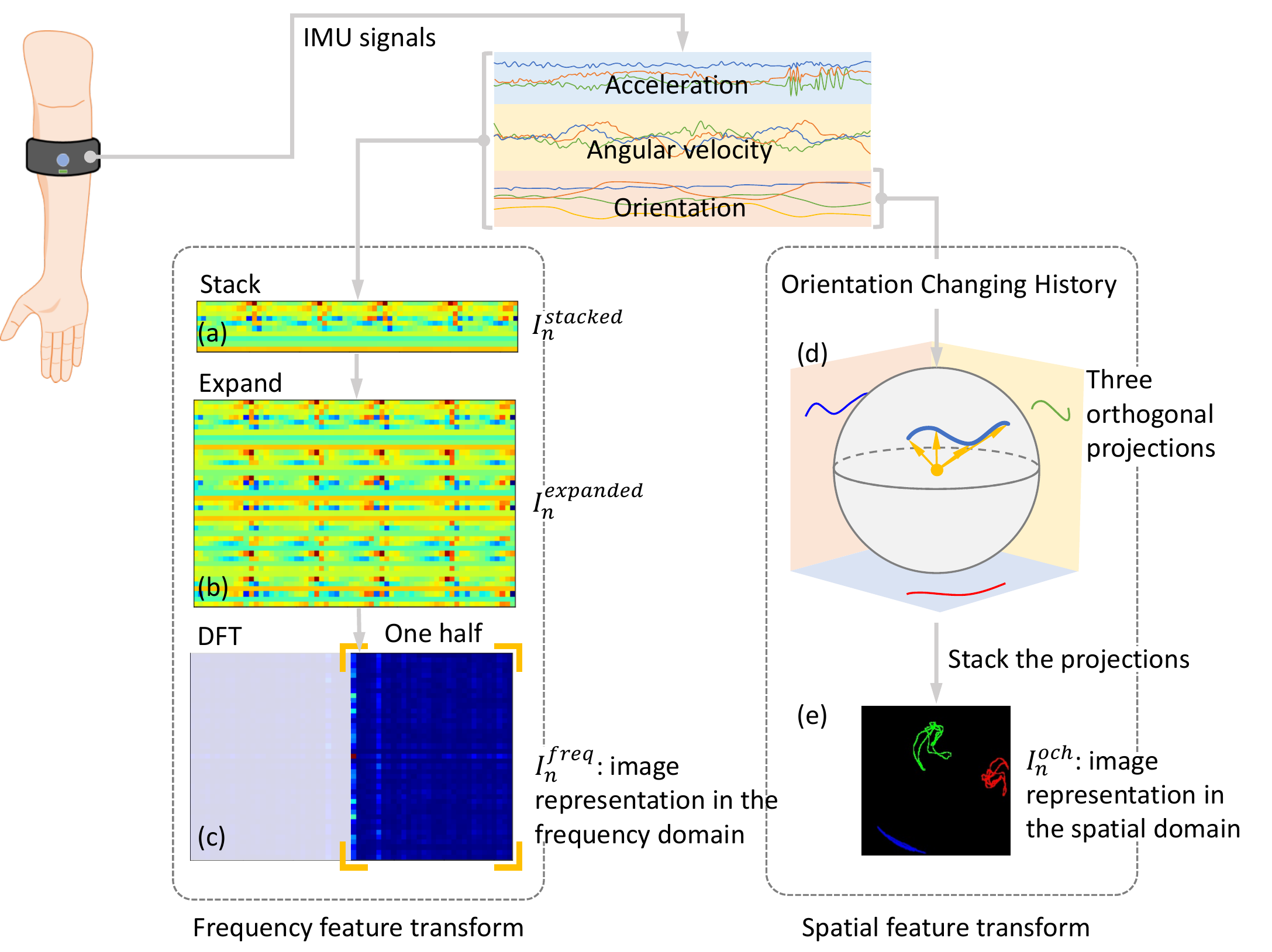}
\end{center}
%\vspace*{-15pt}
   \caption{Illustration of the feature transforms for wearable sensor signals.}
%\label{fig:long}
\label{fig:signal_preprocessing}
\vspace{0pt}\end{figure}

%CNN can be applied to 1D signals as well.
To take advantage of the powerful learning ability of CNNs on images, we propose to transfer the time-series IMU sensor signals to the image representation. As shown in Figure~\ref{fig:signal_preprocessing}, the frequency feature transform assembles the sensor signals in a special pattern such that the hidden correlations among different channels of sensor signals are revealed; and the spatial feature transform uncovers the changing history of orientation signals in the spatial domain. Both feature transform mechanisms enable a CNN model to learn the most discriminative features from images, which are not possible in the original time-series sensor signals.

\textbf{Frequency feature transform}: 
%\textbf{Frequency Domain} 
Frequency domain analysis is a commonly used technique for signal pattern recognition. 
%In this modality, we assemble signals into 2D images and apply frequency domain analysis on it. One important motivation of doing so is to discover the correlation among neighboring signals. The frequency feature transform process is described as
%\begin{equation}
%I_n^{freq}=\mathcal{T}_{freq}(s_n^{})
%\end{equation}
%where $\mathcal{T}_{freq}$ denotes a series of operations for feature transform and $I_n^{freq}$ is the resulted image.
%
Rather than directly applying the frequency transform to time-series signals, we propose a new way to unveil the hidden correlations among sensor signals: 
1) The 10-channel signals $s_n$ in an IMU sample are stacked row by row as an image $I_n^{stacked}$ with the size of $10\times64$ (Fig.~\ref{fig:signal_preprocessing}(a));
2) We expand the 10-row image with a shuffling algorithm~\cite{jiang2015human} to form $I_n^{expanded}$ (Fig.~\ref{fig:signal_preprocessing}(b)) with the size of $42\times64$. The idea here is to make every pair of 10 channels have the chance to be row-neighbors in the image, then the correlations among different channels can be exposed and be further detected by a CNN model;
3) Two-dimensional (2D) Discrete Fourier Transform (DFT) is applied to $I_n^{expanded}$ to get the representation in the frequency domain to analyze the frequency characteristics. Only its logarithmic magnitude is taken to form the image $I_n^{freq}$ (Fig.~\ref{fig:signal_preprocessing}(c));
4) Due to the conjugate symmetry of Fourier Transforms
\begin{equation}
\begin{cases}
\begin{split}
I_n^{freq}(u, v)=I_n^{freq}(-u, -v)\, ,\\
I_n^{freq}(-u, v)=I_n^{freq}(u, -v)\, ,
\end{split}
\end{cases}
\end{equation}
where $u$ and $v$ represent the two directions of an image, we can use only a half to represent the DFT image to remove the redundancy.
This will reduce the architectural complexity and the number of training parameters for the CNN model. Here we keep using the notation $I_n^{freq}$ to represent the one-half (the first and fourth quadrants) of DFT image for simplicity.

% OCH plotter
\textbf{Spatial feature transform}: 
%\textbf{Spatial Domain} 
Implementing feature transform in the frequency domain unavoidably abandons the spatial information from the signals, which motivates us to introduce the second mechanism to exploit the spatial information included in the raw signals. 
%For the spatial feature transform, 
Since recovering the spatial trajectory from IMU data is not an easy task, here we develop an {\it orientation changing history (och)} image to represent the pose-changing information of the subject in the spatial domain
\begin{equation}
I_n^{och}=\mathcal{T}_{och}(q_n^{})
\end{equation}
where $\mathcal{T}_{och}$ is the spatial feature transform and $I_n^{och}$ is the resulted image.
In the spatial feature transform described below, only the orientation information $q_n$ is considered. 

First, a unit vector $\vec{v}_{ref}=[0, 0, 1]$ is rotated by $q_n$ to generate a direction vector $\vec{v}_{n, t}$ by
\begin{equation}
\vec{v}_{n,t}=\vec{q}_{n,t} * \vec{v}_{ref}
\label{eq:vector}
\end{equation}
where $*$ denotes the rotation operation defined as
\begin{equation}
\vec{q}*\vec{v} = [(\vec{q}\otimes [v^x, v^y, v^z, 0])\otimes \vec{q}^*]_{1:3}
\end{equation}
where $\otimes$ is the quaternion multiplication, defined as 
\begin{equation}
\vec{q}_1 \otimes \vec{q}_2=
\begin{bmatrix}
q_1^w q_2^x + q_1^x q_2^w + q_1^y q_2^z - q_1^z q_2^y\\
q_1^w q_2^y + q_1^y q_2^w + q_1^z q_2^x - q_1^x q_2^z\\
q_1^w q_2^z + q_1^z q_2^w + q_1^x q_2^y - q_1^y q_2^x\\
q_1^w q_2^w - q_1^x q_2^x - q_1^y q_2^y - q_1^z q_2^z
\end{bmatrix}^T
\end{equation}
where $\vec{q_1}=[q_1^x, q_1^y, q_1^z, q_1^w]$, $\vec{q_2}=[q_2^x, q_2^y, q_2^z, q_2^w]$, and 
$\vec{q}^*$ is the conjugate of $\vec{q}$:
\begin{equation}
\vec{q}^*=[-q^x, -q^y, -q^z, q^w].
\end{equation}

\noindent Then, the orientation changing history can be represented by a series of orientation vectors at different time steps. 
\begin{equation}
\mathbf{v}_n^{och}=[v_{n,1}^{}, v_{n,2}^{}, \dots, v_{n,t}^{}], t\in[1, T]
\end{equation}
which is essentially a set of points on the unit sphere surface. 

Secondly, these points are projected onto three orthogonal planes (Fig.~\ref{fig:signal_preprocessing}(d)). On each plane, the points are connected with line segments sequentially to form orientation changing curves in an image. 

\begin{figure}[h]
\begin{center}
\includegraphics[width=.5\linewidth]{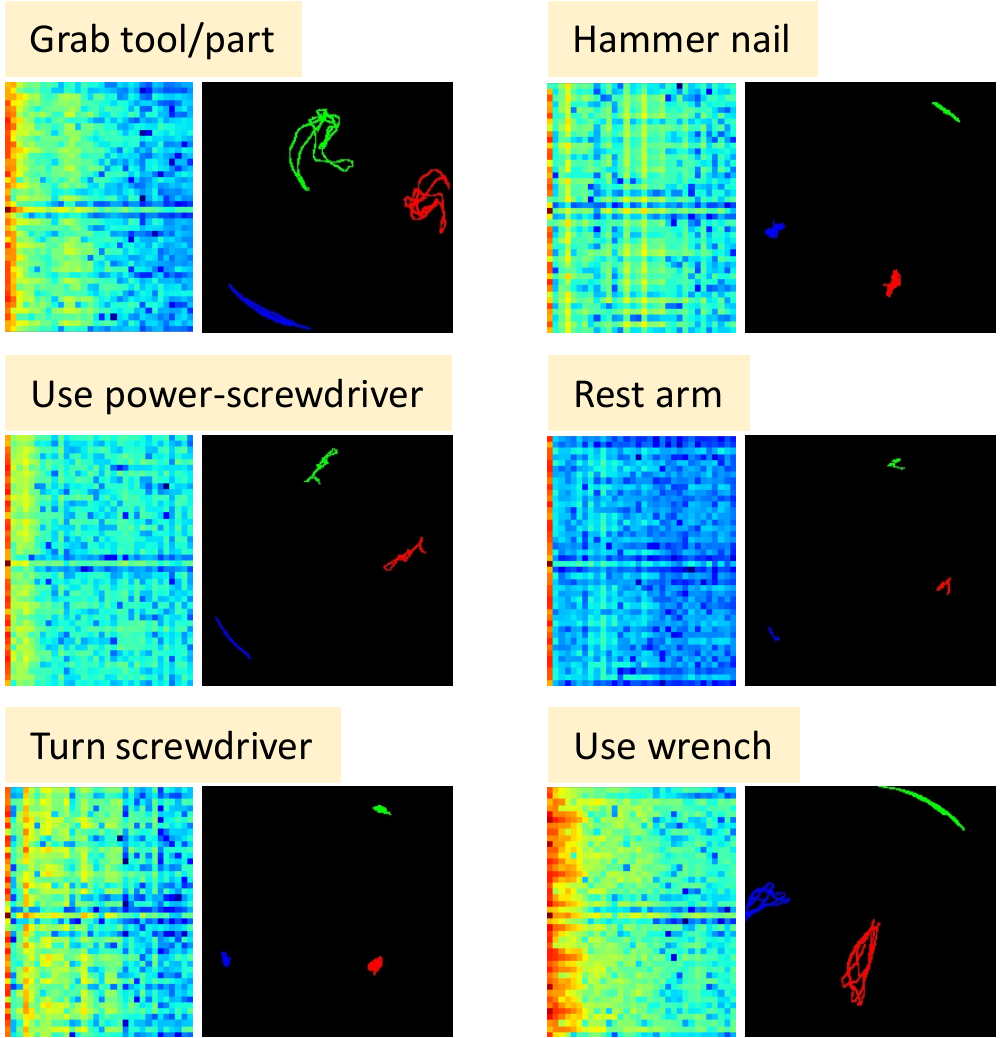}
\end{center}
%\vspace*{-15pt}
   \caption{Examples of IMU image representations by the frequency and spatial feature transforms.}
%\label{fig:long}
\label{fig:activity_image}
\vspace{0pt}\end{figure}

Finally, these three projected images are stacked as a 3-channel image $I_n^{och}$ which is represented in red, green and blue color, respectively (Fig.~\ref{fig:signal_preprocessing}(e)).
Figure~\ref{fig:activity_image} shows some examples of image representations in the frequency and spatial domain, from one subject on six activities, from which we can observe unique patterns of each activity.

\subsection{Visual Signal Representation}
Besides the two mechanisms of feature transforms on the IMU sensor signals, since the recorded video contains rich visual contexts of the worker's activity and visual-based activity recognition also has shown promising results~\cite{ji20133d, carreira2017quo}, we introduce two other mechanisms to represent the video at two levels.

\textbf{Frame-level visual representation}: At the frame level, the middle frame of a video clip is selected as an image representation of an activity, which focuses on the worker's static posture and surrounding environment. The operation is denoted as 
\begin{equation}
I_n^{frame}=\mathcal{T}_{frame}(v_n^{})
\end{equation}
where $\mathcal{T}_{frame}$ is the operation to extract the middle frame from a video clip $v_n$.

\textbf{Video-level visual representation}: At the video-clip level, the video clip samples are sampled again to make each video clip have the defined length of frames for a CNN model (to be described in Section~\ref{multimodal_arch}). The operation is
\begin{equation}
V_n^{clip}=\mathcal{T}_{clip}(v_n^{})
\end{equation}
where $V_n$ is the resulted fixed-length video-clip from the operation of $\mathcal{T}_{clip}$.

% mean on emg
%For a sEMG sample, the 8-channel signals are averaged along each channel, forming an 8-dimensional vector, which represents the level of muscle activation. 

% finally
%Finally, we have $N$ activity images $X_{i}^{IMU}$ and $N$ sEMG vectors $X_i^{sEMG}$, where $i\in[1,N]$.

\subsection{Kinematics-based Data Augmentation}
%\paragraph{Data Augmentation}
For deep learning approaches, a large amount of labeled data are needed to train a valid model with decent performance of generalization. Nevertheless, it is always time-consuming and costly to collect such big data with labels annotated. Data augmentation that synthesizes additional data derived from original ones, is a commonly-used technique to resolve the data shortage problem.
%CNNs require large amount of data to learn millions of parameters. But collecting big data is time consuming. Data augmentation generates additional samples derived from original samples for training.
% Traditional augmentation
Traditionally,  image data augmentation techniques include implementing a series of image transformation operations, such as rotating, scaling, shifting, flipping, shearing, etc., on the original images, to generate more image data. 
The image transformation can introduce more variations and still keep the recognizable contents,
and thus it is applied to generate more data for images and video-clips from the visual signal representation.
However, the variations introduced by the basic image transformation is not physically-realistic in our sensor signal context. 

To include more reasonable variations in the training dataset, we propose a kinematics-based augmentation method to generate more wearable sensor signal samples, rather than implementing image data augmentation on those images resulted from feature transforms.
More specifically, the kinematics-based augmentation refers to creating variations by spatial rotation and mirroring on the four channels of orientation signals.

Suppose we have a four-channel orientation signal represented as a quaternion $\vec{q}_{n,t}$, a new orientation $\hat{\vec{q}}$ can be generated by rotating $\vec{q}_{n,t}$ with
%\begin{equation}
%\vec{q}=[q^x, q^y, q^z, q^w]
%\end{equation}
\begin{equation}
\hat{\vec{q}}=\vec{r}_a^\theta \otimes \vec{q}_{n,t}
\end{equation}
where $\vec{r}_a^\theta$ represents a rotation quaternion of an angle $\theta$ about an axis $\vec{a}=[a^x, a^y, a^z]$. It can be calculated by
%$\vec{r}=[r^x, r^y, r^z, r^w]$ and $\vec{q}=[q^x, q^y, q^z, q^w]$
%
%\begin{equation}
%\hat{\vec{q}}=[\hat{q}^x, \hat{q}^y, \hat{q}^z, \hat{q}^w]=
%\begin{bmatrix}
%r^w q^x + r^x q^w + r^y q^z - r^z q^y\\
%r^w q^y + r^y q^w + r^z q^x - r^x q^z\\
%r^w q^z + r^z q^w + r^x q^y - r^y q^x\\
%r^w q^w - r^x q^x - r^y q^y - r^z q^z
%\end{bmatrix}^T
%\end{equation}
%
\begin{equation}
\vec{r}_a^\theta=[a^x\sin{(\theta/2)}, a^y\sin{(\theta/2)}, a^z\sin{(\theta/2)}, \cos{(\theta/2)}].
\end{equation}

Applying mirroring to the original data is to add variations in some situations, for example, the armband is worn in different dominant arms for different subjects.
%\begin{equation}
%\vec{v}' = \vec{q}*\vec{v}_{ref}
%\end{equation}
First, the vector $\vec{v}^{mirror}$ mirrored from the current direction vector $\vec{v}_{n,t}$ (Eq.~\ref{eq:vector}) against a certain plane can be calculated with
\begin{equation}
\vec{v}^{mirror} = \vec{v}_{n,t} - 2\vec{v}_{n,t} \cdot \vec{n}^T \cdot \vec{n}
\end{equation}
where $\vec{n}$ is the normal vector of the given plane. 

Then the mirrored quaternion $\bar{\vec{q}}=[\bar{q}^x, \bar{q}^y, \bar{q}^z, \bar{q}^w]$, representing the transition between the two vectors $v_{ref}$ and $v_{mirror}$, can be obtained by
\begin{equation}
\begin{split}
[\bar{q}^x, \bar{q}^y, \bar{q}^z] = \vec{v_{ref}}\times \vec{v^{mirror}} \\
\bar{q}^w = 1 + \vec{v_{ref}}\cdot \vec{v^{mirror}}
\end{split}
\end{equation}
where $\times$ and $\cdot$ are the cross and dot products, respectively.

For the other six channels of linear acceleration and angular velocity, since their measurements are relative to the sensor's coordinate systems, rotation and mirror operation do not affect the values. Some random noises (uniformly distributed in the range of $\pm5\%$ of the original signals) are added to simulate the possible fluctuations.

\section{Multi-modal Recognition}\label{multimodal_arch}
In this section the developed multi-modal approach for worker activity recognition is detailed: four deep learning architectures created for different input modalities are presented; the cost function for training each modality is introduced; and the inference fusion strategies to output the recognition result are described.

\subsection{Deep Learning Architectures of Four Input Modalities}
After the preprocessing, signal representation generation and data augmentation described in Section~\ref{signals_preprocessing}, there are $N$\footnote{Here we use the same notation for simplicity but this $N$ is larger than the one in Eq.~\ref{eq:Dn} due to the data augmentation.} data samples $\{X_{1}, \cdots, X_{N}\}$, each of which contains four different inputs:
\begin{equation}
%\begin{split}
X_n \\
= \{I_n^{freq}, I_n^{och}, I_n^{frame}, V_n^{clip}\}, n\in[1,N]
%\end{split}
\end{equation}
where $I_n^{freq}, I_n^{och}, I_n^{frame}$ and $V_n^{clip}$ are the four inputs of frequency feature transform, spatial orientation changing history (och) feature transform, frame-level visual representation and video-level visual representation, respectively. 

%% Unused equations
%\begin{equation}
%%\begin{split}
%S(i, j)\\
%= (I*K)(i,j)\\
%=\sum_m \sum_n{I(m, n)K(i-m, j-n)}
%%\end{split}
%\end{equation}

%\begin{equation}
%a_j^{(l+1)}\\
%= \sigma(b_j^l+\sum_{k=1}^{n_f^i}W_{jk}^l*a_k^l)
%\end{equation}
%where $a_j^{(l+1)}$ denotes the feature map $j$ in layer $l+1$, $\sigma$ , $n_f^l$ is the number of convolutional filters in layer l. The weights of the convolution filter and the bias vector are denoted by $W_{jk}^l$ and $b_j^l$.
%ReLU [] is chosen as the activation function $\sigma$.

For the three image inputs, $I_n^{freq}, I_n^{och}$ and $I_n^{frame}$, 2D convolutional operation~\cite{ji20133d} is applied to extract features layer by layer. The value at position $(x,y)$ in the $j$th feature map of the $i$th layer is computed by
\begin{equation}
v_{ij}^{xy}=g\Bigg(b_{ij}+\sum_k\sum_{p=0}^{P_i-1}\sum_{q=0}^{Q_i-1}w_{ijk}^{pq}v_{(i-1)k}^{(x+p)(y+q)}\Bigg)
\end{equation}
where $g(\cdot)$ denotes a non-linear activation function. $b_{ij}$ is the bias for this feature map, $k$ is the index of the feature maps in layer $(i-1)$, $w_{ijk}^{pq}$ is the value at the position $(p,q)$ of the kernel connected to the $k$th feature map, and $P_i$ and $Q_i$ are the height and width of the two-dimensional kernel, respectively.

%% 3D CNN
For video-clip input $V_n^{clip}$, 3D convolutional operation~\cite{ji20133d} is applied to deal with the additional temporal dimension. The value at position $(x,y,z)$ in the $j$th feature map of the $i$th layer is given by
\begin{equation}
v_{ij}^{xyz}=g\Bigg(b_{ij}+\sum_k\sum_{p=0}^{P_i-1}\sum_{q=0}^{Q_i-1}\sum_{r=0}^{R_i-1}w_{ijk}^{pqr}v_{(i-1)k}^{(x+p)(y+q)(z+r)}\Bigg)
\end{equation}
where $R_i$ is the size of the 3D kernel along the temporal dimension, $w_{ijk}^{pqr}$ is the $(p,q,r)$th value of the kernel connected to the $k$th feature map in the previous layer.

%\subsection{Classifiers}
The feature maps obtained from a series of convolutional operations are flattened as a feature vector. To solve the classification problem, the vector is further input to a multi-layer neural network. 
The value of the $j$th neuron in the $i$th fully connected layer, denoted as $v_{ij}$, is given by
\begin{equation}
v_{ij}=g\Bigg(b_{ij}+\sum_{k=0}^{K_{(i-1)}-1}{w_{ijk}v_{(i-1)k}}\Bigg),
\end{equation}
where $b_{ij}$ is the bias term, $k$ indexes the set of neurons in the $(i-1)$th layer connected to the current feature vector, $w_{ijk}$ is the weight value in the $i$th layer connecting the $j$th neuron to the $k$th neuron in the previous layer. 
%A activation function $g(\cdot)$ can be applied to introduce the nonlinearity.
%The layers can be stacked to form a complex architecture.

\begin{figure}[h]
\begin{center}
\includegraphics[width=1.\linewidth]{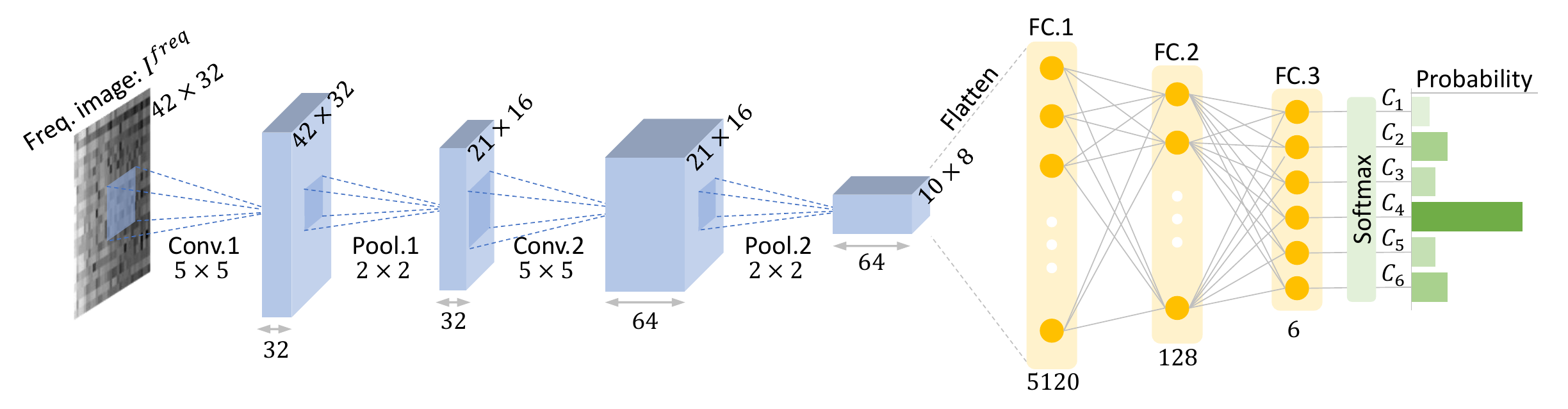}
\end{center}

   \caption{The architecture of our CNN model for $I_n^{freq}$. \lq Conv.\rq\ and \lq Pool.\rq\ denote the operations of convolution and pooling, respectively.}
%\label{fig:long}
\label{fig:cnn_architecture}
\end{figure}

In details, the proposed CNN models for the four input modalities are described as follows:

$I_n^{freq}$: The architecture of our CNN model for $I_n^{freq}$ is illustrated in Figure~\ref{fig:cnn_architecture}. It accepts the frequency image as the input, and outputs a probability distribution of the 6 activities.
%\begin{figure*}[h]\vspace*{4pt}
%%\centerline{\includegraphics{fx1}\hspace*{5mm}\includegraphics{fx1}}
%\centerline{\includegraphics{figures/cnn_architecture}}
%\caption{(a) first picture; (b) second picture.}
%\vspace*{22pt}\end{figure*}
$I_{n}^{freq}$ has the size of $42\times32\times1$ (height, width, depth, respectively) and is normalized to the interval $[0,1]$ before being fed into two $5\times5$ convolutional layers for feature extraction. Each convolutional layer is down-sampled to a half by implementing a $2\times2$ max pooling layer.
% fc
%Then the feature map from the second pooling layer having the size of $10\times8\times64$ is flattened into a $5120$-dimension feature vector, which is subsequently densified by a fully connected layer to a 6 dimensional feature vector.
%
The classification module accepts the $10\times8\times64$ feature map from the last pooling layer and flattens it as a $5120$ feature vector.
Then, two fully connected layers are used to densify the feature vector to the dimensions of 128 and $C$ sequentially, where $
C$ is the number of worker activity classes. Finally, this $C$-dimensional score vector $S ([S_1, ..., S_c, ..., S_C])$ is transformed to output the predicted probabilities with a softmax function as follows:
\begin{equation}
P(y_n=c|X_n)=\frac{\exp(S_c)}{\sum^C_{c=1}\exp(S_c)}
\end{equation}
where $P(y_n=c|X_n)$ is the predicted probability of being class $c$ for sample $X_n$.

%The rectified linear unit or ReLU, which is defined by the activation function $g(z)=\max\{0,z\}$, is used.

$I_n^{och}$ and $I_n^{frame}$
: For these two input modalities, 
%since $I_n^{och}$ and $I_n^{frame}$ have more common contents, 
we use transfer learning to solve the image classification problem instead of building and training CNN models from scratch.
%to extract image features: VGG architecture description
%VGG is the abbreviation of Visual Geometry Group (a lab at University of Oxford). 
%They publish their VGG network (a CNN network) which has been trained on a huge dataset (including 1000 classes) and achieved good accuracy.
To extract image features, we use a VGG network~\cite{simonyan2014very} pretrained on the ImageNet dataset~\cite{imagenet_cvpr09}. For each image input, the feature vector obtained from the fully connected layer FC7 in the VGG model is used to represent the image, then a new classifier is designed on top of it to output the prediction on activity class.

%Here we use the pre-trained VGG network to transfer its learning ability into our area to solve our problem. 

$V_n^{clip}$: The video-clip input $V_n^{clip}$ contains spatial-temporal information. We use the C3D model pretrained on the Sports-1M dataset~\cite{tran2015learning, jia2014caffe,karpathy2014large}. The C3D network reads sequential frames and outputs a fixed-length feature vector every 16 frames. We extract activation vectors from the fully connected layer FC6-1, which is then connected to a new classifier to predict the worker activity class.

%Dropout has been proved to be a powerful regularization technique used to avoid the overfitting, which randomly drops units from the neural network during training~\cite{srivastava2014dropout}. Therefore, it is implemented after each pooling layer in our CNN model. 

\subsection{Training}
Training a deep learning model refers to a process of optimizing the network's weights $w$ to minimize a chosen loss function using training data $X$. The commonly used regularized cross entropy~\cite{Goodfellow-et-al-2016} is chosen as the loss function:
\begin{equation}
\mathcal{L}(w) = \sum^N_{n=1} \sum^C_{c=1}y_{nc}\log[P(y_n=c | X_n)]+\lambda l_2(w)
\end{equation}

\noindent where $y_{nc}$ equals to 0 when the ground truth label of an input image $X_n$ is the $c$th label, and equals to 1 otherwise. To penalize large weights during training, an $l_2$ regularization term is applied to the loss function, where $\lambda$ is its coefficient.
To mitigate overfitting issues, the dropout regularization~\cite{srivastava2014dropout} is used during training, which randomly drops neuron units from the neural network.

%and the dropout rate is a hyperparameter that needs to be tuned. 

% tf.reduce\_mean(cross entropy) + 1e-5*regularizer
%Optimization, Adam, SGD, Initialization,

%For the backpropagation process:
%\begin{equation}
%\frac{\partial \mathcal{L}}{\partial w} = \frac{\partial s}{\partial w}
%\end{equation}

\subsection{Inference Fusion}

%Due to the high inter-class similarities, some activities may not be correctly recognized from certain modalities. The inference relying on only one modality may not be convincing enough. Therefore, we propose a multi-modal inference fusion strategy in order to augment the speculation of each individual modality.
Just like how human uses five senses to perceive the world, multi-modal approach has the opportunity to integrate all the information and make a comprehensive understanding of the learning problem.  %prediction. The integration of multiple modalities can have a comprehensive understanding of the learning problem.
Mathematically, each individual model can return a probability distribution on the worker activity prediction, we can design different strategies to fuse the inferences from different models: 

\textbf{Maximum fusion.}  This method reports the maximum output within a list of predictions.
\begin{equation}
S_c^{max} = \max_{m\in\{1,2,\dots,M\}}{p_c^m}
\end{equation}
where $m$ is the index of different models and $M$ is the total number of models.

\textbf{Average fusion.} In this method, we adopt the average to fuse the outputs of different modalities, i.e., 
\begin{equation}
S_c^{avg} = \frac{1}{M}\sum_{m=1}^M{p_c^m}
\end{equation}

\textbf{Weighted fusion.}
%In the inference fusion step, predictions from all individual views are fused together. 
We introduce the informativity value $\gamma^m$ to evaluate the prediction confidence of each modality $m$. $\gamma^m$ is calculated with Eq.~\ref{eq:informativity},
which is modified from the Shannon entropy of a discrete probability distribution to vary in the interval of $[0,1]$.
\begin{equation}\label{eq:informativity}
\gamma^m=\frac{\sum^K_{k=1}p_k^m\log{p_k^m}}{\log{K}}+1
\end{equation}
where $m$ is the index of modalities and $k$ is the index of top-$K$ candidates. $p_k^m$ represents the probability of the $k$th class candidate at the $m$th model. $I^m$ will be close to 0 if all the top-$K$ candidates have similar probabilities (i.e., $p_k^m\approx 1/K$), and 1 if the probability of top-$1$ class candidate is about reaching 1 (i.e., $p_1^m\approx 1$).

Then every predicted probability $p_k^m$ of the $m$th model is weighted by $\gamma^m$ of this model and the weighted maximum fusion and the weighted average fusion scores are
\begin{equation}
S_c^{max-w} = \max_{m\in\{1,2,\dots,M\}}{\gamma^mp_c^m}
\end{equation}

\begin{equation}
S_c^{avg-w} = \frac{1}{M}\sum_{m=1}^M{\gamma^mp_c^m}
\end{equation}

For the above four fusion strategies, the final predicted label is chosen as the one that maximizes the fusion score (e.g., for weighted average fusion, $c^*=\arg \max_c S_c^{avg-w}$).

\section{Experiments and Evaluation Metrics}\label{experiment}

\subsection{Implementation Details}\label{implementation}
The CNN architectures of the four input modalities described in the previous sections are constructed using TensorFlow~\cite{tensorflow2015-whitepaper} and Keras libraries.
They are trained individually so that each of them can make its own inference for further decision fusion. The SGD optimizer is used in training, with the momentum of 0.9, the learning rate of 0.001 and the regularizer coefficient of 1e-5. The batch size for each of the four models is 512, 64, 64 and 512, respectively, which is limited by the computation memory. 
%dropout rate,  0.5
%and the dropout rates are hyperparameters which need to be tuned during training. 
The number of training epochs is 1000 and 100 for the first modality $I_n^{freq}$ and the other modalities, respectively.
We use a workstation with one 12-core Intel Xeon processor, 64GB of RAM and two Nvidia Geforce 1080 Ti graphic cards for the training jobs. It takes approximately 30 minutes to train each model for a leave-one-out experiment.

%%TODO: Talk about preprocessing for c3d and vgg here.

\subsection{Evaluation Metric}
Two evaluation policies are conducted, i.e., half-half and leave-one-out policies. In the half-half evaluation, after randomly shuffling, one half of the dataset is used for training and the other half is kept for testing. In the leave-one-out evaluation, the samples from 7 out of 8 subjects are used for training, and the samples of the left one subject are reserved for testing.
We employ several commonly used metrics~\cite{Goodfellow-et-al-2016} to evaluate the classification performance, which are listed as follows:
\begin{itemize}
\item Accuracy
\begin{equation}
Accuracy=\frac{\sum^N_n1(\hat{y}_n=y_n)}{N}
\end{equation}
\item Precision and Recall
\begin{equation}
\begin{split}
Precision=\frac{TP}{TP+FP} \\
Recall=\frac{TP}{TP+FN}
\end{split}
\end{equation}
\item $F_1$ score
\begin{equation}
F_1=2\cdot\frac{Precision\cdot Recall}{Precision+Recall}
\end{equation}
\end{itemize}
where $1()$ is an indicator function. For a certain class $y_i$, True Positive (TP) is defined as a sample of class $y_i$ that is correctly classified as $y_i$; 
%True Negative (TN) means a sample from a class other than $y_i$ is correctly classified as \lq not $y_i$\rq; 
False Positive (FP) means a sample from a class other than $y_i$ is misclassified as $y_i$; False Negative (FN) means a sample from the class $y_i$ is misclassified as another \lq not $y_i$\rq\ class. $F_1$ score is the harmonic mean of Precision and Recall, which ranges in the interval [0,1].

%\textbf{Training}

\section{Results and Discussion}\label{result}
In this section, we first perform evaluations of the data augmentation methods.
Then, we compare the performance of different fusion methods.
After that, we explore various modalities and their combinations for an ablation study. 
The performance of our approach on some public dataset is also reported.
Then, we conduct visualizations for a better understanding of the CNN model.
Finally, future research needs are discussed.

\subsection{Evaluation of the Data Augmentation Methods}
To evaluate the effectiveness of our proposed kinematics-based augmentation (KA)  method, we compare it to the jittering augmentation (JA) method~\cite{sermanet2011traffic}, which has been proved to be an effective method and is commonly used in CNN-based image classification tasks.

%We use different $N_{APS}$ (number of augmentations per sample) when generate augmented samples from an original one.
% how PA works
For the KA method, four rotation angles $\{\pm \pi/8, \pm \pi/4\}$ are selected for rotation augmentation, i.e., new samples are generated by implementing rotation on the original signal samples, and two mirroring planes  $yz$-plane and $xz$-plane are chosen for mirroring augmentation, overall yielding 6 augmented samples for each actual sample.
Then the amount of the augmented training dataset is 6 times more than the original one.
Note that the augmentation is applied directly on each original signal sample $s_n$ before the feature transforms.

% how JA works
As for the JA method, to have a fair comparison with the KA method, 6 augmented samples are generated by randomly translating in the range of $\pm10\%$ of the image width/height, scaling in the range of $[0.9, 1.1]$ ratio, and rotating in the range of $[-5, +5]$ degrees.
In the JA method, the augmentation is applied to each image $I_n^{freq}$ and $I_n^{och}$ after the feature transforms.

% how JA+KA works
We also evaluate the performance of the JA+KA method, in which the augmented data from the JA and KA methods are integrated.
The leave-one-out evaluations of the two modalities $I_n^{freq}$ and $I_n^{och}$ on our activity dataset with the different augmentation methods are shown in Table~\ref{tab:aug} (the half-half accuracies are not considered for the comparison purpose because they are about reaching 100\%).
All the three augmentation methods have accuracy improvements compared with the models without using data augmentation.
% m1 observation
For $I_n^{freq}$, the JA method improves the accuracy from 88.0\% to 88.7\%, and the KA method outperforms the JA method, whose accuracy is 90.2\%. By combining the JA and KA methods, the accuracy is slightly further improved to 90.5\%. 
% m2 observation
For $I_n^{och}$, the accuracy is improved from 63.6\% to 65.0\% with the JA method, and is further improved by using the KA method, which is 77.3\%, 12 percentage points higher than the JA method. However, the JA+KA method does not further improve the accuracy and its accuracy 75.3\% is lower than the KA method.

\begin{table}[] \footnotesize
\caption{Comparison (\%) of accuracy regarding to data augmentation: None (without data augmentation), JA (jittering augmentation), KA (kinematics augmentation) and JA+KA, for the leave-one-out experiments.}
\begin{center}
\begin{tabular}{ccccc}
\hline
 \multirow{2}{*}{Modalities} &  \multicolumn{4}{c}{Data Augmentation Methods}\\
\cline{2-5}
 & None & JA & KA & JA+KA \\ \hline
$I_n^{freq}$ & 88.01 & 88.71 & 90.18 & \textbf{90.51} \\
$I_n^{och}$ & 63.59 & 65.01 & \textbf{77.27} & 75.33 \\
\hline 
\end{tabular}
\end{center}
\label{tab:aug}
\end{table}

%Therefore, $N_{APS}=18$ is chosen for comparison with state-of-the-art methods in the following sub-section.

Overall, the data augmentation techniques, JA and KA, demonstrate the effectiveness in improving the model performance, because the augmentation process introduces more variations to the training dataset to simulate the potential variations in the unseen samples, which pushes the deep learning model to learn the most discriminative features and makes the training more robust.
%It shows that data augmentation is very effective in improving the ability of generalization. 
Meanwhile, the KA method outperforms the JA method. It is because rather than introducing variations to the image, like what JA method does, KA method directly generates some physically-realistic variations to the original signal sample, which is more effective to augment the dataset to be more comprehensive.
%% Our choice
Although JA+KA method improves the performance of $I_n^{freq}$ slightly compared with KA method, it does not for $I_n^{och}$. Because JA+KA method has a larger amount (i.e., 2 times) of training data and $I_n^{och}$ has a more complex architecture than $I_n^{freq}$, which makes the training less efficient, we choose KA method for both of the modalities in the following study as a compromise between performance and training efficiency.

%Why the performance does not continue increasing as the increase of the $N_{APS}$ because the new augmented views are already fidelity and cannot provide more information for the CNN model.
%depth image has more information than a grayscale image, although they both have one color channel. 

\subsection{Evaluation of Different Fusion Methods}
%% how
Each of the four input modalities generates a vector output before the fusion step. Then, these vector outputs are fused to have only one score vector as the final output.
To study the effect of the four different fusion methods: 1) maximum fusion, 2) average fusion, 3) weighted maximum fusion and 4) weighted average fusion,
a set of experiments are conducted on our activity dataset.

%% observation
The comparisons of the fusion performance, in terms of accuracy, precision, recall and F score, are listed in Table~\ref{tab:fusion}. The average fusion method performs better than the maximum fusion method for all the metric items. The weighted maximum method has the same accuracy as the maximum method but lower precision, recall and F score. The weighted average method has lower performance than the average method.
The two weighted methods do not contribute additional improvement as they did in~\cite{tao2018american}.
%% our choice
%The average fusion has the highest performance, i.e., 97.2\%, 97.0\%, 96.8\% and 96.8\%, respectively. 
Therefore, the average method is chosen as the fusion strategy for our following experiments.

\begin{table}[] \footnotesize
\caption{Comparison (\%) of different fusion methods for the leave-one-out experiments.}
\begin{center}
\begin{tabular}{ccccc}
\hline
Fusion Methods & Accuracy & Precision & Recall & F Score \\ \hline
Maximum     & 93.68 & 92.48 & 92.50 & 91.09      \\
Average      & \textbf{97.17} & \textbf{97.04} & \textbf{96.82} & \textbf{96.81}      \\
Weighted Max.     & 93.68 & 92.45 & 92.49 & 91.07        \\
Weighted Avg.     & 96.79 & 96.38 & 96.28 & 96.04      \\
\hline 
\end{tabular}
\end{center}
\label{tab:fusion}
\end{table}

\subsection{Evaluation of Different Input Modalities}
A central idea of our approach is that the reasoning based on multiple modalities can significantly improve the inference performance based on single modality.
To validate this idea, we perform a comprehensive ablation study where we progressively increase the number of modalities and try different modality combinations. 
The performance of these cases in terms of accuracy, precision, recall and F score with two evaluation policies (half-half and leave-one-out) is summarized in Table~\ref{tab:result}.

To simplify the abbreviation, we use $\mathcal{M}_1$, $\mathcal{M}_2$, $\mathcal{M}_3$ and $\mathcal{M}_4$ to represent the four input modalities, $I_n^{freq}$, $I_n^{och}$, $I_n^{frame}$ and $V_n^{clip}$, respectively.
For the single-modal cases, although $\mathcal{M}_1$ and $\mathcal{M}_2$ are both based on the IMU signals, $\mathcal{M}_2$ shows lower performance because it only uses the 4 orientation channels out of the 10 channels. 
Also, it demonstrates that the frequency feature transform provides more discriminative features for activity recognition.
% m3 and m4
$\mathcal{M}_3$ performs better than $\mathcal{M}_4$, which shows that the current pretrained VGG model can extract more discriminative features than the C3D model.
%because a 2D CNN model is easier to train and achieve high performance than a 3D CNN model.
% highest
Overall, $\mathcal{M}_1$ achieves the highest performance in the single-modal cases, whose metric items are accuracy (90.2\%), precision (90.7\%), recall (89.5\%) and F score (87.6\%), respectively.

For the dual-modal cases, all the 6 combinations are evaluated. All the cases have better results compared with their related single-modal cases, e.g., $\mathcal{M}_{\{2, 3\}}$ performs better than both $\mathcal{M}_2$ and $\mathcal{M}_3$. 
% highest
$\mathcal{M}_{\{1, 3\}}$ has the highest accuracy as their individual modalities are also the highest two for the single-modal cases.

For the triple-modal cases, 4 combinations are tested. The fusion of more modalities further improve the performance than the duel-modal cases. 
% highest
$\mathcal{M}_{\{1, 3, 4\}}$ has the highest accuracy as their individual modalities are also the highest three for the single-modal cases.

Finally, a quad-modal case $\mathcal{M}_{\{1, 2, 3, 4\}}$ including all the four modalities is experimented, which achieves the highest performance.
%Overall, by implementing the multi-modal fusion, more activities are correctly recognized. $\mathcal{M}_{\{1, 2, 3, 4\}}$ demonstrates the highest performance in all the cases.
Therefore, we choose the quad-modal architecture for our model. 

%In the above three multi-modal cases, we use average fusion to fuse the inferences from different input modalities.

\begin{table}[h]\centering \footnotesize
\caption{Overall performance (\%) of the half-half (hh) and leave-one-out (loo) experiments.}
\begin{tabular}{ccccccccc}
\hline
\multirow{2}{*}{Methods}  & \multicolumn{2}{c}{Accuracy}    &\multicolumn{2}{c}{Precision}    &\multicolumn{2}{c}{Recall}    & \multicolumn{2}{c}{F Score}    \\ \cline{2-9}
  & hh    & loo    & hh    & loo    & hh    & loo    & hh    & loo    \\
\hline
Previous~\cite{tao2018worker} & 97.6 & 87.4 & 97.8 & 89.0 & 97.5 & 89.5 & 97.7 & 87.6 \\ \hline
$\mathcal{M}_1$  & 99.5   & 90.2 & 99.5 & 90.7 &  99.6 & 90.9 & 99.5 & 90.3     \\
$\mathcal{M}_2$  & 93.0   & 77.3 & 92.3 & 77.5 & 92.5 & 78.3 & 92.4 & 75.0 \\
$\mathcal{M}_3$  & 100   & 86.8 & 100 & 83.0 & 100 & 83.2 & 100 & 81.3 \\
$\mathcal{M}_4$  & 100   & 80.8 & 100 & 79.1 & 100 & 77.7 & 100 & 74.3 \\ 
\hline
$\mathcal{M}_{\{1,2\}}$ & 99.6 & 91.1 & 99.6 & 91.5 & 99.6 & 92.1 & 99.6 & 91.4 \\
$\mathcal{M}_{\{1,3\}}$ & 100 & 94.8 & 100 & 94.9 & 100 & 94.6 & 100 & 94.3 \\
$\mathcal{M}_{\{1,4\}}$ & 100 & 92.2 & 100 & 93.1 & 100 & 91.3 & 100 & 90.1 \\
$\mathcal{M}_{\{2,3\}}$ & 100 & 90.3 & 100 & 90.9 & 100 & 87.8 & 100 & 87.0 \\
$\mathcal{M}_{\{2,4\}}$ & 100 & 85.0 & 100 & 84.0 & 100 & 82.8 & 100 & 80.2 \\
$\mathcal{M}_{\{3,4\}}$ & 100 & 89.5 & 100 & 86.3 & 100 & 86.0 & 100 & 84.2 \\ 
\hline
$\mathcal{M}_{\{1,2,3\}}$ & 100 & 95.3  & 100 & 95.4 & 100 & 95.4 & 100 & 95.2 \\
$\mathcal{M}_{\{1,2,4\}}$ & 100 & 93.9  & 100 & 93.5 & 100 & 94.1 & 100 & 93.0 \\
$\mathcal{M}_{\{1,3,4\}}$ & 100 & 95.9  & 100 & 95.2 & 100 & 94.8 & 100 & 94.2 \\
$\mathcal{M}_{\{2,3,4\}}$ & 100 & 92.6  & 100 & 90.4 & 100 & 90.7 & 100 & 89.5 \\ 
\hline
$\mathcal{M}_{\{1,2,3,4\}}$   & \bf{100}   & \bf{97.2} & \bf{100} & \bf{97.0} & \bf{100} & \bf{96.8} & \bf{100} & \bf{96.8} \\ 
\hline
\end{tabular}
\begin{tablenotes}\footnotesize
\item[*] $\mathcal{M}_1$, $\mathcal{M}_2$, $\mathcal{M}_3$ and $\mathcal{M}_4$ represent the four input modalities, $I_n^{freq}$, $I_n^{och}$, $I_n^{frame}$ and $V_n^{clip}$, respectively.
$\mathcal{M}_{\{\cdot, \cdot\}}$  denotes a multi-modality model, e.g., $\mathcal{M}_{\{2, 3\}}$ represents the fusion of the $\mathcal{M}_2$ and $\mathcal{M}_3$ modalities. We use average fusion method to fuse multiple modalities together.
\end{tablenotes}
\label{tab:result}
\end{table}

% half-half
For the half-half experiments, almost all of the testing samples are correctly recognized. It is higher than the leave-one-out experiments. This is because all the testing subjects are seen in the half-half experiment, while the testing subject in the leave-one-out experiment is unseen.

\subsection{Performance Comparison on the Public Dataset}
To validate the generalization of our method, a commonly-used public dataset for human activity recognition, PAMAP2 dataset~\cite{reiss2012introducing}, is also chosen for comparison. This dataset has 12 human activities (lying, sitting, standing, walking, running, cycling, Nordic walking, ascending stairs, descending stairs, vacuum cleaning, ironing and rope jumping) 
captured by three IMU sensors (worn on the wrist, chest and ankle, respectively), and
%which is comparable to our WA6 dataset. 
the activities are performed by 9 different subjects.
% and have approximately 10k samples in total. 
%Examples of the depth image of the 10 signs in this dataset are shown in Figure~\ref{fig:dataset_digit_raw}. Each image contains background and the hand region is not cropped or annotated. 
% reason
Since the PAMAP2 dataset does not include video recordings, we evaluate the performance of our CNN models of the first two modalities on it. 
% implementation
%In the experiments, we use subject 6 for testing and the rest subjects for training.
%result
The performance comparison of several existing deep learning models on the PAMAP2 dataset is listed in Table~\ref{tab:comparison_pamap2}. 
%
%Here we use the same evaluation protocol as the one that is used in the listed literature.
%
Using the same evaluation protocol,
our model achieves the best recognition accuracy, 94.2\%, compared with other methods in the literature.
%which outperforms the results reported in the literatures. 
% compared with Jiang
%The MVA+IF method further improve the leave-one-out accuracy to 100\%, which means that the  inference fusion strategy successfully classify the left 0.3\% samples that are misclassfied using only MVA.

%\begin{table}[h]\footnotesize
%\begin{center}
%\begin{tabular}{lc}
%\hline
%Method & Accuracy\\
%\hline
%Ronao and Cho. (2015a)~\cite{ronao2015deep} & 94.79\\
%Ronao and Cho (2015b)~\cite{ronao2015evaluation} & 90.00 \\
%Jiang and Yin (2015)~\cite{jiang2015human} & 95.18\\
%Ronao and Cho (2016)~\cite{ronao2016human} & 94.61 \\
%\hline
%Our model & \bf{95.89} \\
%\hline
%\end{tabular}
%\end{center}
%\caption{Performance (\%) comparison of existing deep models on the PAMAP2 activity dataset.}
%\label{tab:comparison_pamap2}
%\end{table}

\begin{table}[h]\footnotesize
\begin{center}
\begin{tabular}{lc}
\hline
Method & Accuracy\\
\hline
Hammerla et al. (2016)~\cite{hammerla2016deep} & 93.70\\
Murahari et al. (2018)~\cite{murahari2018attention} & 87.50 \\
Zeng et al. (2018)~\cite{zeng2018understanding} & 89.96\\
Xi et al. (2018)~\cite{xi2018deep} & 93.50 \\
Xu et al. (2019)~\cite{xu2019innohar} & 93.50 \\
\hline
Our model & \bf{94.16} \\
\hline
\end{tabular}
\end{center}
\caption{Performance (\%) comparison of existing deep models on the PAMAP2 activity dataset.}
\label{tab:comparison_pamap2}
\end{table}

\subsection{Visualizing the Class Activation Map of $\mathcal{M}_3$}

\begin{figure*}[h]
\begin{center}
\includegraphics[width=1.\linewidth]{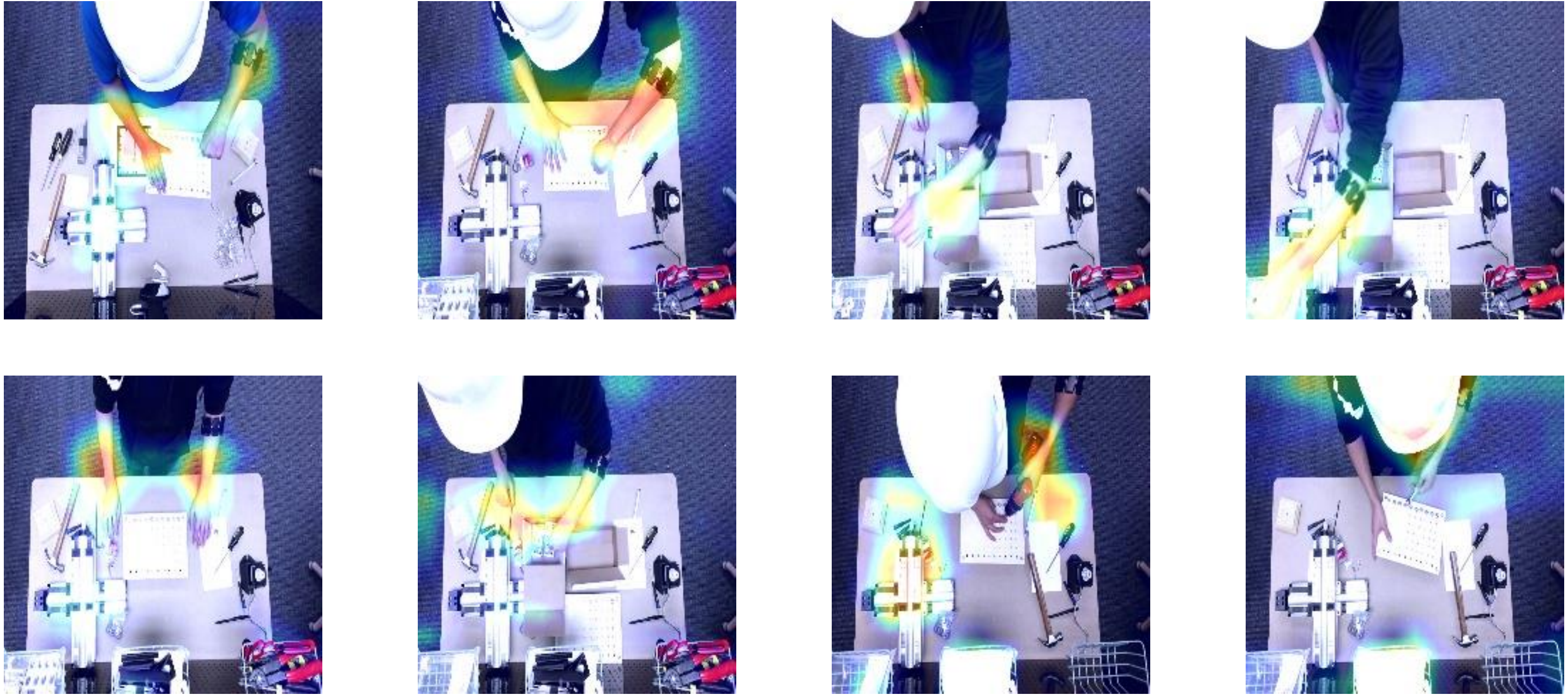}
\end{center}
\caption{Examples of Class Activation Map (CAM) Visualization.}
%\label{fig:long}
\label{fig:cam}
\end{figure*}

It is known that a deep learning model driven by a large amount of data can achieve superior performance for various tasks, such as the image classification task for our third modality $\mathcal{M}_3$. However, it is usually treated as a black box and criticized for weak interpretability, due to the high model complexity and tremendous hidden weights. 
To have an intuitive interpretation of the connections between image contents and the predicted activity class, we visualize the class activation map (CAM),
which is a score heatmap associated with a specific activity class, computed for every location of an input image, representing how strong the connection is between each location and a specific activity class. 
A set of CAM visualizations are shown in Figure~\ref{fig:cam}, where the generated heapmaps are overlaid onto the input images. We can see that the model is able to focus on the hand and tool regions, where exactly the interaction happens.
%which visualizes heatmaps of class activation on top of input images. 

\subsection{Future Research Needs}
At present, we conduct the multi-modal recognition of 6 basic activities. 
To further push the current approach to the practical application, some directions for future work are considered, such as recruiting more subjects to learn more working styles, optimizing data augmentation techniques to add more variations to the collected data, and exploring different methods of signal preprocessing and feature extraction to fully exploit the recorded signals. In addition, more fusion methods can be explored and every modality can be further improved to reach their optimal performance.

\section{Conclusion}\label{conclusion}
Worker behavior awareness is crucial towards human-centered intelligent manufacturing. In this paper, we proposed a multi-modal approach for worker activity recognition. Two sensors (wearable device and camera) were adopted to perceive the worker, and four modalities were built to recognize the activity independently. Then, inference fusion was implemented to achieve an optimal understanding of the worker's behavior.
%using the Inertial Measurement Unit (IMU) signals obtained from a Myo armband and videos from a Kinect camera. 
% generalize
%The main point of this approach is that, benefits can be derived from multiple modalities. 

% new idea
We designed two novel mechanisms to produce image representations of the IMU sensor signals in both the frequency and spatial domains.
% new idea
A kinematics-based data augmentation method was developed to generate more physically-realistic variations in the training dataset. This performs better than the traditional data augmentation method.
% dataset: 
A worker activity dataset has been established, which currently involves 8 subjects and contains 6 common activities in assembly tasks (i.e., grab a tool/part, hammer a nail, use a power-screwdriver, rest arms, turn a screwdriver and use a wrench).
The multi-modal approach is evaluated on the dataset and achieves 100\% and 97\% recognition accuracy in the half-half and leave-one-out experiments, respectively.
Our approach can be further generalized to other sensors, modalities, and working contexts.

\section*{Acknowledgment}
This research work is supported by the National Science Foundation grants CMMI-1646162 and NRI-1830479, and also by the Intelligent Systems Center at Missouri University of Science and Technology. Any opinions, findings, and conclusions or recommendations expressed in this material are those of the authors and do not necessarily reflect the views of the National Science Foundation.

%\section*{References}
%\bibliography{references}

\end{document}